\documentclass{article}

\usepackage[utf8]{inputenc} 
\usepackage[T1]{fontenc}    
\usepackage{hyperref}       
\usepackage{url}            
\usepackage{booktabs}       
\usepackage{amsfonts}       
\usepackage{nicefrac}       
\usepackage{microtype}      
\usepackage{xcolor}         

\usepackage{amsmath}
\usepackage{graphicx}
\usepackage{wrapfig}
 \usepackage{array,multirow,graphicx}
\usepackage{enumitem}

\title{Data Augmentation for Compositional Data: \\ Advancing Predictive Models of the Microbiome}

\author{%
  Elliott Gordon-Rodriguez 
    \\
  Columbia University\\
  \texttt{eg2912@columbia.edu} \\
  \and 
    Thomas P. Quinn \\
  Independent Scientist\\
  \texttt{contacttomquinn@gmail.com} \\
  \and 
    John P. Cunninghham \\
  Columbia University\\
  \texttt{jpc2181@columbia.edu} \\
}

\begin{document}

\maketitle

\begin{abstract}
  Data augmentation plays a key role in modern machine learning pipelines.
  While numerous augmentation strategies have been studied in the context of computer vision and natural language processing, less is known for other data modalities.
  Our work extends the success of data augmentation to \emph{compositional data}, i.e., simplex-valued data, which is of particular interest in the context of the human microbiome.
  Drawing on key principles from compositional data analysis, such as the \emph{Aitchison geometry of the simplex} and \emph{subcompositions}, we define novel augmentation strategies for this data modality.
  Incorporating our data augmentations into standard supervised learning pipelines results in consistent performance gains across a wide range of standard benchmark datasets.
  In particular, we set a new state-of-the-art for key disease prediction tasks including colorectal cancer, type 2 diabetes, and Crohn's disease.
  In addition, our data augmentations enable us to define a novel contrastive learning model, which improves on previous representation learning approaches for microbiome compositional data.
Our code is available at \url{https://github.com/cunningham-lab/AugCoDa}.
\end{abstract}

\section{Introduction}

Data augmentation, i.e., generating synthetic training examples to be used in model fitting, 
is a core component of modern deep learning pipelines \cite{shorten2019survey, feng2021survey}.
In computer vision, augmentations such as image resizing and shifting have been used since as early as LeNet-5 \cite{lecun1998gradient}.
These and many other augmentations have become essential to highly successful state-of-the-art architectures, ranging from AlexNet \cite{krizhevsky2012imagenet} and ResNet \cite{he2016deep} to recent contrastive learning models such as SimCLR \cite{chen2020simple} and MoCo \cite{he2020momentum}.
As such, a growing body of literature has emerged to develop and characterize data augmentation techniques, particularly in computer vision \cite{perez2017effectiveness, zhang2017mixup, devries2017improved, cubuk2018autoaugment, cubuk2020randaugment, yun2019cutmix}, as well as natural language processing \cite{sennrich2015improving, devlin2018bert, feng2021survey}.
However, defining useful data augmentations is highly domain-dependent, and fewer works have studied augmentations for more general data modalities, such as tabular data \cite{zhang2017mixup, yoon2020vime}.
The goal of our work is to extend the success of data augmentation to a previously unexplored data modality; namely, {compositional data}.

%

Compositional data (CoDa) are those that represent the parts of a whole, and therefore carry only \emph{relative} information.
Equivalently, we can think of CoDa as a set of datapoints living in the simplex:
\begin{align}
\mathcal D = \left\{ \boldsymbol x_i \in \Delta^{p-1} \right\}_{i=1}^n, \ \ \mathrm{ where } \ \ \  \Delta^{p-1} = \left\{ \boldsymbol x \in \mathbb R_+^{p} : \sum_{j=1}^{p} x_j = 1 \right\}.
\end{align}
Examples of CoDa arise throughout the sciences, most notably in microbiology \cite{gloor2016compositional, gloor2017microbiome, quinn2018understanding, avalos2018representation}, geochemistry \cite{buccianti2006compositional, buccianti2014compositional}, ecology \cite{douma2019analysing}, materials science \cite{na2014compositional}, economics \cite{fry2000compositional}, 
and machine learning \cite{gordon2020continuous, gordon2020uses, gordon2022normalizing}. 
As a result, starting with the seminal work of Aitchison \cite{aitchison1982statistical}, numerous bespoke techniques have been developed for the statistical analysis of CoDa \cite{pawlowsky2007lecture, pawlowsky2011compositional, filzmoser2018applied}.

A rapidly growing area of application for CoDa is the human microbiome, which comprises the populations of microorganisms that reside inside and on the human body \cite{turnbaugh2007human}.
Microbiome data arise from an inexhaustive sampling procedure as a result of high-throughput sequencing \cite{gloor2016compositional, quinn2021critique}.
In particular, each feature typically represents the \emph{relative abundance} of some species of microorganism; as such, each observation can be normalized to the simplex prior to downstream analyses  \cite{gloor2016compositional, gloor2017microbiome, quinn2018understanding, lo2019metann, oh2020deepmicro}.

The microorganisms that constitute the microbiome are known to impact human physiology, both in health and in disease \cite{gilbert2018current, methe2012framework, gevers2014treatment}. 
Thus, a central problem in CoDa is to learn the association $\bold x_i \mapsto y_i$, where $\bold x_i \in \Delta^{p-1}$ denotes the microbial composition, and $y_i$ denotes the disease status of the $i$th subject in a clinical study.
For example, the composition of the gut microbiome, as recorded from a stool sample, can be predictive of colorectal cancer, which is the third most prevalent form of cancer \cite{favoriti2016worldwide, kostic2012genomic}.
This observation offers the potential for a noninvasive alternative to traditional colonoscopy procedures used for early detection of colorectal cancer \cite{tarallo2019altered}.
In turn, accurate predictive models for microbiome CoDa are a key stepping stone towards achieving this potential \cite{mulenga2021feature, kostic2012genomic}.
Note that cancer is just one application; the demand for improved predictive models holds more broadly across medicine and biological science \cite{kostic2012genomic, gilbert2018current}.
For example, the microbiome has been linked to type 2 diabetes, Crohn's disease, obesity, and others \cite{hartstra2015insights, wright2015recent, qin2012metagenome, gevers2014treatment}.

Classical techniques including logistic regression, support vector machines, and random forests have been widely used as predictive models for microbiome data \cite{vangay2019microbiome, cammarota2020gut, liu2021practical}.
More recently, specialized deep learning architectures such as DeepCoDa \cite{quinn2020deepcoda},  MetaNN \cite{lo2019metann}, and DeepMicro \cite{oh2020deepmicro} have been developed.
However, the capacity of these deep networks and other expressive models has been limited by the low sample size and high dimensionality of typical microbiome studies. 
These characteristics have also spurred the use of strong regularization through early stopping, weight decay and dropout, among others \cite{topccuouglu2020framework, quinn2020deepcoda, lo2019metann, oh2020deepmicro}.
However, no previous works have explored the use of data augmentation for CoDa,\footnote{With the possible exception of dropout, which can be viewed as a form of data augmentation. Note that applications of dropout to CoDa, such as \cite{lo2019metann}, use standard implementations that do not exploit the special structure of CoDa.} which, as we shall demonstrate, provides a cheap and simple technique for boosting the performance of predictive models for this data modality.

%

Careful consideration of the sample space will motivate our novel data augmentation strategies for CoDa.
Our work draws on foundational principles from the field of CoDa, such as the \emph{Aitchison geometry} of the simplex \cite{aitchison1982statistical}, which we combine with popular techniques from the literature on data augmentation, such as {Mixup} \cite{zhang2017mixup} and CutMix \cite{yun2019cutmix}.
This combination will lead us to define custom data augmentation strategies for CoDa, such as \emph{Aitchison Mixup} and \emph{Compositional CutMix}.
In turn, incorporating these novel augmentations into existing supervised learning pipelines will result in consistent performance gains across a wide range of microbiome datasets, including a dozen standard benchmarks from the Microbiome Learning Repo \cite{vangay2019microbiome}.
These performance gains are particularly large for some deep models, for example, DeepCoDa gains more than 10\% in test AUC for discriminating colorectal cancer from healthy controls, and over 20\% for type 2 diabetes.
The gains are also significant across other expressive model families, including random forests and gradient boosting machines.
Overall, 
we set a new state-of-the-art on 8 out of 12 benchmark learning tasks, including clinically relevant disease prediction tasks.
Of the remaining 4 tasks, 2 were easily separable, with 100\% test accuracy irrespective of whether data augmentation was used.
Importantly, our augmentations rarely hurt model performance, and in the few instances that this was the case, the drops were typically of only 1\% in test AUC.

In addition to supervised learning, our novel augmentations will allow us to define \emph{contrastive representation learning for CoDa}, 
which to the best of our knowledge, represents the first contrastive learning model for compositional data.
Our novel data augmentations are at the core of this approach; the contrastive loss uses randomly augmented training examples to define a self-supervised optimization objective whose labels are generated from unlabelled data \cite{hadsell2006dimensionality, dosovitskiy2014discriminative, chen2020simple, he2020momentum}.
In particular, our contrastive model is trained to discriminate between compositions that were generated as random augmentations of the same training example, and those that were generated as random augmentations of different training examples.
Our implementation is adapted from SimCLR \cite{chen2020simple}, but using a smaller network architecture together with our novel augmentations.
Unlike SimCLR, we use our individual augmentation strategies in isolation, and find them sufficient to surpass previous representation learning approaches for microbiome CoDa \cite{avalos2018representation, oh2020deepmicro}.
Altogether, we expect our data augmentations will enable significant future progress, possibly in combination with novel architectures, in both supervised and representation learning pipelines for microbiome CoDa.


\section{Related Work} \label{sec:related}


\textbf{Data augmentation:} numerous data augmentation strategies have been proposed in 
 the context of image data \cite{shorten2019survey} and text data \cite{feng2021survey}.
LeNet-5 \cite{lecun1998gradient} used random shifts and resizing, and AlexNet \cite{krizhevsky2012imagenet} used reflections and color distortions, which have become widely adopted in computer vision \cite{he2016deep}.
More general purpose data augmentations include {Mixup}, which generates synthetic samples by taking convex combinations of training examples, and has been applied to both image and tabular data \cite{zhang2017mixup}.
Random masking, whereby random input features are hidden during training, has also seen a diverse range of applications, including natural language processing (e.g., BERT \cite{devlin2018bert}), computer vision (e.g., Cutout \cite{devries2017improved}), and tabular data (e.g., VIME \cite{yoon2020vime}).
In the context of CoDa, masking is related to \emph{subcompositions}, 
which we will use to define an analogous augmentation on the simplex.
CutMix \cite{yun2019cutmix} is akin to both masking and Mixup, in that random patches from different images are pasted together to generate new synthetic images.
We note that a subsequent line of work has developed techniques to automatically select optimal data augmentations for a given dataset \cite{cubuk2018autoaugment, cubuk2020randaugment}, somewhat akin to hyperparameter optimization.
While the data augmentation strategies that we present in the next Section could also be combined and finetuned in similar ways, such experimentation falls outside the scope of our work and is left to future research.



\textbf{Microbiome models:} 
random forests remain a strong baseline across many microbiome studies, due to their expressivity and robustness \cite{topccuouglu2020framework}. 
Gradient boosting machines such as XGBoost provide similarly strong performance and have also enjoyed significant adoption \cite{topccuouglu2020framework}.
AutoML \cite{he2021automl} has shown potential for microbiome data, with a recent approach called mAML achieving state-of-the-art results on several benchmarks \cite{yang2020maml}. mAML uses cross-validation to automatically select the best model class and hyperparameters for each learning task.
In addition, several specialized deep learning architectures have been proposed for microbiome CoDa.
DeepCoDa \cite{quinn2020deepcoda} introduced the {log-bottleneck} layer, which ensures the hidden units remain {scale-invariant}, a key desiderata in CoDa models.
This architecture obtained strong predictive performance using weight decay regularization.
MetaNN \cite{lo2019metann} employs a multilayer architecture regularized using dropout \cite{srivastava2014dropout}.
The authors of MetaNN also considered generating synthetic data from a negative binomial distribution, but found no performance improvements beyond the use of dropout.
DeepMicro \cite{oh2020deepmicro} uses a deep autoencoder architecture to learn low-dimensional representations of the microbial composition.
In turn, these representations are fed to a downstream classifier trained with a supervised objective.


\textbf{Contrastive learning:} 
the goal of contrastive learning is to learn low-dimensional representations by optimizing some \emph{pretext task}, where the objective function is similar to those used for supervised learning, but using labels that are derived from unlabelled data only \cite{hadsell2006dimensionality}.
Commonly, the pretext task is to discriminate augmented instances of the same training example from augmented instances of different training examples \cite{dosovitskiy2014discriminative, chen2020simple, he2020momentum}.
Recent contrastive learning architectures have enjoyed tremendous success, setting the state-of-the-art across a range of computer vision benchmarks \cite{chen2020simple, he2020momentum, khosla2020supervised, caron2020unsupervised}.
We highlight SimCLR \cite{chen2020simple} for its relative simplicity and strong empirical performance.
This method uses a nonlinear projection head between the representations and the contrastive loss computed with normalized embeddings.
Our own experiments on contrastive learning borrow these elements from SimCLR, but use a smaller network architecture together with our specialized data augmentations for CoDa.



\section{Methods} \label{sec:methods}

In the following subsections, we introduce 3 novel data augmentation techniques for CoDa. 
Note that many variations of our augmentation schemes could be considered; the versions we present here are intended to be as concise as possible. 
Our goal is not to design augmentations that are empirically or theoretically optimal, but rather to demonstrate the effectiveness of our novel methodology by establishing simple and performant baselines.
For instance, whenever a random mixing parameter is required, we use a $U(0,1)$ distribution, even though other choices can likely result in increased performance.
For clarity, we will focus on the classification setting; generalizing to regression problems is straightforward.
We will use the notation $\bold v, \bold x \in \Delta^{p-1}$ for simplex-valued vectors, $\lambda \in \mathbb R$ a scalar, and $\mathcal D = \{ \bold x_i, y_i \}$ our training data.
\begin{wrapfigure}{r}{0.45\textwidth}
  \begin{center}
  \vspace{-2mm}
    \includegraphics[width=0.36\textwidth]{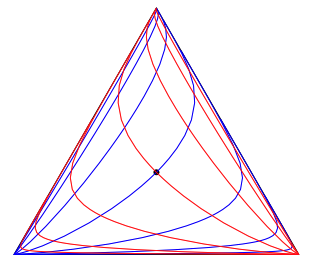} 
  \end{center}
  \caption{Orthogonal grid on $\Delta^2$, in the Aitchison sense   \cite{pawlowsky2007lecture}.
  The centroid of the simplex corresponds to the additive identity.
  The red lines are parallel and equally spaced by 1 unit in Aitchison distance, as are the blue lines.
  The red lines are also orthogonal to the blue lines.
 } \label{fig:aitch} 
 \vspace{-32mm}
\end{wrapfigure}

\subsection{Aitchison Mixup}

Aitchison \cite{aitchison1982statistical} defined a Hilbert space structure on the simplex, known as the \emph{Aitchison geometry}, with the following vector addition, scalar multiplication, and inner product:
\begin{align}
\bold v \oplus {\bold x} &= \frac{1}{\sum_{j=1}^p v_j x_j} [v_1  x_1, \dots, v_p  x_p],
\\
\lambda \odot {\bold x} &= \frac{1}{\sum_{j=1}^p x_j^\lambda} [x_1^\lambda, \dots, x_p^\lambda],
\\ \nonumber
\langle \bold v , {\bold x} \rangle  &= \frac{1}{2p} \sum_{j=1}^p \sum_{k=1}^p \log \left( \frac{v_j}{v_{k}} \right) \log \left( \frac{ x_j}{ x_{k}} \right).
\end{align}
This geometry provides a principled definition of linear combinations and a distance metric on the simplex; parallel and orthogonal lines are illustrated in Figure \ref{fig:aitch}.
The Aitchison geometry is closely related to the \emph{centered-log-ratio} transformation \cite{aitchison1982statistical}, which is an inverse of the softmax and defines an isometry between $\Delta^{p-1}$ and Euclidean space \cite{pawlowsky2007lecture}.
Taken together, these properties form the basis of much of CoDa methodology \cite{aitchison1982statistical, pawlowsky2007lecture, pawlowsky2011compositional, filzmoser2018applied}.

Our first data augmentation strategy, which we call \emph{Aitchison Mixup}, is based on taking linear combinations of the training points, in the Aitchison sense.
While general linear combinations may be used, for simplicity we focus on intra-class pairwise convex combinations.
Namely, each new datapoint is sampled as follows:
\begin{enumerate}
 \item Draw a class $c$ from the class prior and draw $\lambda \sim U(0,1)$.
 \item Draw two training points $i_1, i_2$ such that $y_{i_1} = y_{i_2} = c$, uniformly at random.
 \item Set $\bold x^{\text{aug}} = (\lambda \odot \bold x_{i_1}) \oplus ((1 -\lambda) \odot \bold x_{i_2})$ and $y^{\text{aug}} = c$. 
\end{enumerate}
Put more succinctly, for each class we generate convex combinations of the points in that class, in the Aitchison sense.
Note this augmentation strategy is a CoDa analogue of Mixup, which generates Euclidean convex combinations of images and tabular data \cite{zhang2017mixup}.
Much like Mixup is capable of boosting predictive accuracy by regularizing models towards a linear decision boundary in the regions between training examples, Aitchison Mixup aims to enforce the equivalent regularization in the Aitchison geometry of the simplex.

Note that, once we restrict to convex combinations (as opposed to general linear combinations), our augmented data will remain inside the simplex regardless of whether we operate in the Euclidean or the Aitchison geometry.
However, we found the latter to be more empirically effective, perhaps unsurprisingly given the associated vector space structure of the simplex.
Note also that this augmentation strategy may be generalized straightforwardly by taking non-convex combinations, or by including multiple training points from different classes; the study of such approaches is left to future work.

%

\subsection{Random Subcompositions}

In CoDa, a subcomposition of $\bold x \in \Delta^{p-1}$ refers to a lower-dimensional composition, $\bold x^{\text{sub}} \in \Delta^{k-1}$ with $k < p$, formed by taking a subset of the elements of $\bold x$ and renormalizing to a unit total.
Namely, if $j_1, \dots, j_k$ denotes a subset of the indices $\{1, \dots, p\}$, the corresponding subcomposition is defined as:
\begin{align}
 {\bold x^{\text{sub}}} &= \frac{1}{x_{j_1} + \cdots + x_{j_k}} [ x_{j_1}, \dots, x_{j_k}].
\end{align}
We can generate augmented data by taking \emph{Random Subcompositions} of the training points; such a strategy is analogous to masking in language data or cropping in image data.
Since our predictive models require inputs of fixed dimension $p$, rather than discarding elements of $\bold x$ we simply zero them out.
Thus, each new datapoint is generated as follows:
\begin{enumerate}
 \item Draw $\lambda \sim U(0,1)$. Draw a training point $i$ uniformly at random and set $\bold{\tilde x} = \bold x_i$.
 \item For each $j \in \{1, \dots, p\}$, draw $I_j \overset{iid}{\sim} \text{Bernoulli}(\lambda)$, and set $\tilde x_j = 0$ if $I_j = 0$.
 \item Set $\bold x^{\text{aug}} = \bold{\tilde x} / ({\sum_{j=1}^p \tilde x_{j}}) $ and $y^{\text{aug}} = y_i$.
\end{enumerate}
In short, we zero out random entries of the training points and renormalize.
Intuitively, this strategy encourages our predictive models to become robust to partially observed inputs.
Note that, distinctively from random masking, our augmentation includes an additional renormalization to a unit total.
Importantly, this renormalization ensures that the augmented samples remain in the support of the training data, i.e., the simplex.

Note that many CoDa models, including mAML and {DeepCoDa}, apply log transformations to their inputs, and therefore require that these be non-zero.
For this reason, we implement a zero-replacement step where we add a small positive quantity to all the parts of each composition and renormalize.
This small quantity is set to $1/L_i$, where $L_i$ corresponds to the library size from the high-throughput sequencing procedure.
$L_i$ can simply be thought of as a large number such that $1/L_i$ is smaller than the non-zero components of $\bold x_i$.
Note that this transformation is a standard preprocessing step in CoDa \cite{quinn2018understanding}.


\subsection{Compositional CutMix}

Our third augmentation scheme, which we call \emph{Compositional CutMix}, combines elements of our previous 2 augmentations.
Like in Mixup, we generate new datapoints by combining pairs of training points from the same class.
However, instead of combining these training points linearly (in the Aitchison sense), we take complementary subcompositions and renormalize.
Namely, we generate each new datapoint as follows:
\begin{enumerate}
 \item Draw a class $c$ from the class prior and draw $\lambda \sim U(0,1)$.
 \item Draw two training points $i_1, i_2$ such that $y_{i_1} = y_{i_2} = c$, uniformly at random.
 \item For each $j \in \{1, \dots, p\}$, draw $I_j \overset{iid}{\sim} \text{Bernoulli}(\lambda)$, and set $\tilde x_j = x_{i_1j}$ if $I_j = 0$ or $\tilde x_j = x_{i_2j}$ if $I_j = 1$.
 \item Set $\bold x^{\text{aug}} = \bold{\tilde x} / ({\sum_{j=1}^p \tilde x_{j}}) $ and $y^{\text{aug}} = c$.
\end{enumerate}
Note this strategy can be thought of as a CoDa analogue of \emph{CutMix}, whereby new images are formed by pasting together patches from different training images \cite{yun2019cutmix}.
Note also that in computer vision, CutMix and related methods take \emph{local} image patches as opposed to randomly sampled pixels.
Likewise, one could sample microbial subcompositions according to biologically relevant groupings, for example using a phylogenetic tree \cite{quinn2018understanding}.
Such a strategy would require incorporating additional domain knowledge and is left to future work, but we expect it would further increase the quality of our data augmentations.
This remark applies both to Random Subcompositions and Compositional CutMix.

\section{Experiments} \label{sec:exp}

\subsection{Supervised learning}

\begin{table}
  \caption{Evaluation benchmark consisting of 12 binary classification tasks taken from the Microbiome Learning Repo \cite{vangay2019microbiome}, after filtering to datasets containing at least 100 samples with at least 50 in each class.
  For each task we show the number of samples ($n$), the number of features ($p$), a description of the two classes and the number of samples in each, together with a reference to the original studies that each dataset was obtained from. 
  }
  \label{tab:data}
  \centering
  \footnotesize
    \vspace{3mm}
\begin{tabular}{c@{\hspace{1.5\tabcolsep}}c@{\hspace{1.5\tabcolsep}}c@{\hspace{1.5\tabcolsep}}c@{\hspace{1.5\tabcolsep}}c@{\hspace{1.5\tabcolsep}}c@{\hspace{1.5\tabcolsep}}c@{\hspace{1.5\tabcolsep}}c} 
\toprule
Task & $n$ & $p$ & Class 1 / Class 2 & \# in 1 & \# in 2 & Reference \\
\midrule
1 & 140 & 992 & Crohn's disease / Without (ileum) & 78 & 62& \cite{gevers2014treatment} \\
2 & 160 & 992 & Crohn's disease / Without (rectum) & 68 & 92  & \cite{gevers2014treatment} \\
3 & 2070 & 3090 & Gastrointestinal tract / Oral & 227 & 1843 & \cite{methe2012framework} \\
4 & 180 & 3090 & Female / Male & 82 & 98 & \cite{methe2012framework} \\
5 & 404 & 3090 & Stool / Tongue (dorsum) & 204 & 200  & \cite{methe2012framework} \\
6 & 408 & 3090 & Subgingival / Supragingival plaque & 203 & 205 & \cite{methe2012framework} \\
7 & 172 & 980 & Healthy / Colorectal cancer & 86 & 86& \cite{kostic2012genomic} \\
8 & 124 & 2526 & Diabetes / Without &  65  & 59 & \cite{qin2012metagenome} \\
9 & 130 & 2579 & Cirrhosis / Without & 68 & 62  & \cite{qin2014alterations} \\
10 & 199 & 660 & Black / Hispanic & 104 & 95& \cite{ravel2011vaginal}  \\
11 & 342 & 660 & Nugent score high / Low & 97 & 245  &  \cite{ravel2011vaginal}\\
12 & 200 & 660 & Black / White & 104 & 96 & \cite{ravel2011vaginal} \\
\bottomrule
\end{tabular}
\end{table}

We evaluate our augmentation strategies on 12 standard binary classification tasks taken from the Microbiome Learning Repo \cite{vangay2019microbiome}.
These datasets comprise various disease and phenotype prediction tasks, including colorectal cancer \cite{kostic2012genomic}, type 2 diabetes \cite{qin2012metagenome}, Crohn's disease \cite{gevers2014treatment}, and cirrhosis \cite{qin2014alterations}, as well as multiple body sites including the gut, skin, oral cavity, airways, and vagina \cite{methe2012framework, ravel2011vaginal}.
As such, this benchmark provides a comprehensive evaluation for predictive models of the human microbiome \cite{yang2020maml, quinn2020deepcoda, gordon2022learning}.
More details on these 12 learning tasks can be found in Table \ref{tab:data}; note this benchmark is constructed from the original Microbiome Learning Repo by filtering datasets that contain a minimum sample size of 100, with at least 50 in either class.

For each learning task, we take 20 independent 80/20 train/test splits and we fit Random Forest, XGBoost, mAML \cite{yang2020maml}, DeepCoDa \cite{quinn2020deepcoda}, and MetaNN \cite{lo2019metann}, first to the original training data, then on 3 augmented training sets obtained using our 3 augmentation strategies.
Thus, we train a total of $12 \times 20 \times 5 \times 4 = 4\,800$ models.\footnote{We train these models in parallel on a CPU cluster.}
We evaluate test performance using ROC AUC, and we note that our datasets do not present severe class imbalance (with the exception of tasks 3 and 11, both of which were well separated by all our classifiers and therefore had no effect on the overall comparison).

For the augmented training sets, we generated 10 times as many synthetic samples as there were original training examples.
The factor of 10 was chosen so as to obtain a relatively large augmented sample, in order to reduce the sampling variance from our random augmentations.
In turn, we compensate for the fact that our augmented data is then much more numerous than our original training data, by downweighting the synthetic samples by a factor of 10; the total weight of the original and synthetic data is then equal to $1/2$ each.


\begin{table}
  \caption{Data augmentation performance for Aitchison Mixup. We show the test AUC, averaged over 20 train/test bootstraps, for each learning task and predictive model, trained with and without data augmentation. Bold numbers indicate whether the version with or without augmentation performed best. Underlined numbers indicate the overall best model for that task. 
  Models trained with Aitchison Mixup consistently outperformed those without (with the possible exception of MetaNN, which performed worst across the board). Error bars are given in Appendix \ref{app:exp}.
  }
  \label{tab:aitch}
  \centering
  \footnotesize
    \vspace{3mm}
\begin{tabular}{c|c@{\hspace{1\tabcolsep}}c|c@{\hspace{1\tabcolsep}}c|c@{\hspace{1\tabcolsep}}c|c@{\hspace{1\tabcolsep}}c|c@{\hspace{1\tabcolsep}}c}
\toprule
Task & RF & Aug & XGB & Aug & mAML & Aug & DeepCoDa & Aug & MetaNN & Aug \\
\midrule
1 & 0.72 & \textbf{\underline{0.79}} & 0.76 & \textbf{0.79} & 0.72 & \textbf{0.74} & 0.73 & \textbf{\underline{0.79}} & \textbf{0.74} & \textbf{0.74}\\
2 & 0.78 & \textbf{0.82} & \textbf{0.81} & 0.80 & \textbf{0.80} & \textbf{0.80} & 0.78 & \textbf{\underline{0.83}} & \textbf{0.74} & \textbf{0.74}\\
3 & \textbf{1.00} & \textbf{1.00} & \textbf{1.00} & \textbf{1.00} & \textbf{1.00} & \textbf{1.00} & \textbf{1.00} & \textbf{1.00} & \textbf{1.00} & \textbf{1.00}\\
4 & 0.60 & \textbf{\underline{0.64}} & \textbf{0.57} & \textbf{0.57} & 0.56 & \textbf{0.58} & \textbf{0.58} & \textbf{0.58} & 0.50 & \textbf{0.51}\\
5 & \textbf{1.00} & \textbf{1.00} & \textbf{1.00} & \textbf{1.00} & \textbf{1.00} & \textbf{1.00} & \textbf{1.00} & \textbf{1.00} & \textbf{1.00} & \textbf{1.00}\\
6 & 0.81 & \textbf{0.83} & 0.82 & \textbf{0.83} & \textbf{\underline{0.84}} & 0.83 & 0.78 & \textbf{0.82} & 0.75 & \textbf{0.76}\\
7 & \textbf{0.68} & 0.67 & 0.67 & \textbf{0.69} & 0.73 & \textbf{\underline{0.74}} & 0.63 & \textbf{0.73} & \textbf{0.59} & 0.54\\
8 & 0.62 & \textbf{0.65} & 0.66 & \textbf{0.68} & 0.64 & \textbf{0.65} & 0.45 & \textbf{\underline{0.70}} & \textbf{0.64} & \textbf{0.64}\\
9 & \textbf{0.93} & \textbf{0.93} & 0.94 & \textbf{\underline{0.95}} & 0.92 & \textbf{0.93} & 0.84 & \textbf{0.90} & 0.76 & \textbf{0.82}\\
10 & 0.53 & \textbf{0.60} & 0.57 & \textbf{0.61} & 0.61 & \textbf{0.62} & 0.62 & \textbf{\underline{0.63}} & \textbf{0.63} & 0.61\\
11 & \textbf{0.98} & \textbf{0.98} & \textbf{0.98} & \textbf{0.98} & \textbf{0.98} & \textbf{0.98} & \textbf{0.98} & \textbf{0.98} & \textbf{0.96} & 0.95\\
12 & 0.55 & \textbf{0.61} & 0.58 & \textbf{0.65} & \textbf{0.61} & \textbf{0.61} & \textbf{\underline{0.66}} & 0.65 & 0.58 & \textbf{0.60}\\
\midrule
Mean & 0.77 & \textbf{0.79} & 0.78 & \textbf{0.80} & 0.78 & \textbf{0.79} & 0.75 & \textbf{0.80} & \textbf{0.74} & \textbf{0.74}\\
\bottomrule
\end{tabular}
\end{table}

\begin{table}
  \caption{Data augmentation performance for Random Subcompositions, similar to Table \ref{tab:aitch}. Training sets augmented with Random Subcompositions consistently performed better than those without. }
  \label{tab:randsub}
  \centering
  \footnotesize
    \vspace{3mm}
\begin{tabular}{c|c@{\hspace{1\tabcolsep}}c|c@{\hspace{1\tabcolsep}}c|c@{\hspace{1\tabcolsep}}c|c@{\hspace{1\tabcolsep}}c|c@{\hspace{1\tabcolsep}}c}
\toprule
Task & RF & Aug & XGB & Aug & mAML & Aug & DeepCoDa & Aug & MetaNN & Aug \\
\midrule
1 & 0.72 & \textbf{\underline{0.78}} & 0.76 & \textbf{0.77} & \textbf{0.72} & \textbf{0.72} & 0.73 & \textbf{\underline{0.78}} & 0.74 & \textbf{0.76}\\
2 & 0.78 & \textbf{0.81} & \textbf{0.81} & \textbf{0.81} & \textbf{0.80} & \textbf{0.80} & 0.78 & \textbf{\underline{0.85}} & 0.74 & \textbf{0.76}\\
3 & \textbf{1.00} & \textbf{1.00} & \textbf{1.00} & \textbf{1.00} & \textbf{1.00} & \textbf{1.00} & \textbf{1.00} & \textbf{1.00} & \textbf{1.00} & \textbf{1.00}\\
4 & 0.60 & \textbf{\underline{0.62}} & \textbf{0.57} & \textbf{0.57} & \textbf{0.56} & \textbf{0.56} & \textbf{0.58} & 0.52 & \textbf{0.50} & \textbf{0.50}\\
5 & \textbf{1.00} & \textbf{1.00} & \textbf{1.00} & \textbf{1.00} & \textbf{1.00} & \textbf{1.00} & \textbf{1.00} & \textbf{1.00} & \textbf{1.00} & \textbf{1.00}\\
6 & 0.81 & \textbf{0.82} & 0.82 & \textbf{0.83} & \textbf{\underline{0.84}} & 0.83 & 0.78 & \textbf{0.82} & 0.75 & \textbf{0.78}\\
7 & \textbf{0.68} & \textbf{0.68} & \textbf{0.67} & \textbf{0.67} & {0.73} & \textbf{\underline{0.74}} & 0.63 & \textbf{\underline{0.74}} & \textbf{0.59} & 0.53\\
8 & 0.62 & \textbf{0.64} & 0.66 & \textbf{\underline{0.69}} & \textbf{0.64} & 0.63 & 0.45 & \textbf{0.61} & 0.64 & \textbf{0.65}\\
9 & \textbf{0.93} & 0.92 & \textbf{\underline{0.94}} & \textbf{\underline{0.94}} & \textbf{0.92} & \textbf{0.92} & 0.84 & \textbf{0.90} & 0.76 & \textbf{0.82}\\
10 & 0.53 & \textbf{0.58} & 0.57 & \textbf{0.61} & \textbf{0.61} & \textbf{0.61} & \textbf{0.62} & 0.55 & 0.62 & \textbf{\underline{0.63}}\\
11 & \textbf{0.98} & \textbf{0.98} & \textbf{0.98} & \textbf{0.98} & \textbf{0.98} & \textbf{0.98} & 0.98 & \textbf{\underline{0.99}} & \textbf{0.96} & \textbf{0.96}\\
12 & 0.55 & \textbf{0.60} & 0.58 & \textbf{0.63} & \textbf{0.61} & \textbf{0.61} & \textbf{\underline{0.66}} & 0.65 & 0.58 & \textbf{0.59}\\
\midrule
Mean & 0.77 & \textbf{0.79} & 0.78 & \textbf{0.79} & \textbf{0.78} & \textbf{0.78} & 0.75 & \textbf{0.78} & 0.74 & \textbf{0.75}\\
\bottomrule
\end{tabular}
\end{table}

\begin{table}
  \caption{Data augmentation performance for Compositional CutMix, similar to Table \ref{tab:aitch}. Training sets augmented with Compositional CutMix consistently enjoyed better test performance than those without, for all of our predictive models. Note that models trained with this data augmentation set a new state-of-the-art on 8 out of 12 tasks, including disease prediction for colorectal cancer, type 2 diabetes, and Crohn's disease.
 }
  \label{tab:subcomp}
  \centering
  \footnotesize
    \vspace{3mm}
\begin{tabular}{c|c@{\hspace{1\tabcolsep}}c|c@{\hspace{1\tabcolsep}}c|c@{\hspace{1\tabcolsep}}c|c@{\hspace{1\tabcolsep}}c|c@{\hspace{1\tabcolsep}}c}
\toprule
Task & RF & Aug & XGB & Aug & mAML & Aug & DeepCoDa & Aug & MetaNN & Aug \\
\midrule
1 & 0.72 & \textbf{\underline{0.78}} & 0.76 & \textbf{0.77} & 0.72 & \textbf{0.74} & 0.73 & \textbf{0.79} & \textbf{0.74} & \textbf{0.74}\\
2 & 0.78 & \textbf{0.81} & 0.81 & \textbf{0.82} & 0.80 & \textbf{0.81} & 0.78 & \textbf{\underline{0.83}} & 0.74 & \textbf{0.77}\\
3 & \textbf{1.00} & \textbf{1.00} & \textbf{1.00} & \textbf{1.00} & \textbf{1.00} & \textbf{1.00} & \textbf{1.00} & \textbf{1.00} & \textbf{1.00} & \textbf{1.00}\\
4 & 0.60 & \textbf{\underline{0.65}} & 0.57 & \textbf{0.59} & 0.56 & \textbf{0.59} & \textbf{0.58} & 0.57 & \textbf{0.50} & \textbf{0.50}\\
5 & \textbf{1.00} & \textbf{1.00} & \textbf{1.00} & \textbf{1.00} & \textbf{1.00} & \textbf{1.00} & \textbf{1.00} & \textbf{1.00} & \textbf{1.00} & \textbf{1.00}\\
6 & 0.81 & \textbf{0.82} & 0.82 & \textbf{0.83} & \textbf{\underline{0.84}} & \textbf{\underline{0.84}} & 0.78 & \textbf{0.82} & 0.75 & \textbf{0.78}\\
7 & 0.68 & \textbf{0.69} & 0.67 & \textbf{0.68} & \textbf{0.73} & 0.72 & 0.63 & \textbf{\underline{0.76}} & \textbf{0.59} & 0.55\\
8 & 0.62 & \textbf{0.66} & 0.66 & \textbf{\underline{0.72}} & \textbf{0.64} & \textbf{0.64} & 0.45 & \textbf{0.69} & 0.64 & \textbf{0.65}\\
9 & \textbf{0.93} & \textbf{0.93} & 0.94 & \textbf{\underline{0.95}} & 0.92 & \textbf{0.93} & 0.84 & \textbf{0.91} & 0.76 & \textbf{0.81}\\
10 & 0.53 & \textbf{0.57} & 0.57 & \textbf{0.59} & \textbf{0.61} & \textbf{0.61} & \textbf{0.62} & 0.61 & 0.62 & \textbf{\underline{0.65}}\\
11 & \textbf{0.98} & \textbf{0.98} & \textbf{0.98} & \textbf{0.98} & \textbf{0.98} & \textbf{0.98} & 0.98 & \textbf{\underline{0.99}} & \textbf{0.96} & \textbf{0.96}\\
12 & 0.55 & \textbf{0.57} & 0.58 & \textbf{0.63} & \textbf{0.61} & \textbf{0.61} & \textbf{\underline{0.66}} & 0.65 & 0.58 & \textbf{0.61}\\
\midrule
Mean & 0.77 & \textbf{0.79} & 0.78 & \textbf{0.80} & 0.78 & \textbf{0.79} & 0.75 & \textbf{0.80} & 0.74 & \textbf{0.75}\\
\bottomrule
\end{tabular}
\end{table}

Our results are shown in Tables \ref{tab:aitch} (Aitchison Mixup), \ref{tab:randsub} (Random Subcompositions), and \ref{tab:subcomp} (Compositional CutMix): 
\begin{itemize}
 \item Table \ref{tab:aitch} shows that Aitchison Mixup improved the test performance of existing models across a large majority of learning tasks.
On average, we obtained a 5\% gain in test AUC for DeepCoDa, 2\% for random forests and XGBoost, and 1\% for mAML.
MetaNN remained flat, but this was also the worst performing model overall.
Importantly, in the few instances where Aitchison Mixup hurt the performance of a model, the loss was typically of just 1\%.
 \item Table \ref{tab:randsub} shows that Random Subcompositions also result in consistent performance improvements across most models and tasks, though to a slightly lesser degree than Aitchison Mixup. 
 On average, DeepCoDa enjoyed a 3\% gain in test AUC, random forests 2\%, and XGBoost and MetaNN both 1\% (mAML remained flat).
 \item Table \ref{tab:subcomp} shows that Compositional CutMix obtained the strongest classification performance out of our 3 augmentation strategies.
On average, DeepCoDa saw a 5\% gain in test AUC, with 2\% for random forests and XGBoost, and 1\% for mAML and MetaNN.
Moreover, models trained with Compositional CutMix set a new state-of-the-art on 8 out of 12 tasks, including colorectal cancer, type 2 diabetes, and Crohn's disease.
Of the remaining 4 tasks, 3 were tied and only on 1 was the best model one that did not use Compositional CutMix.
 \end{itemize}

\begin{figure} 
  \centering
  \includegraphics[width=\textwidth]{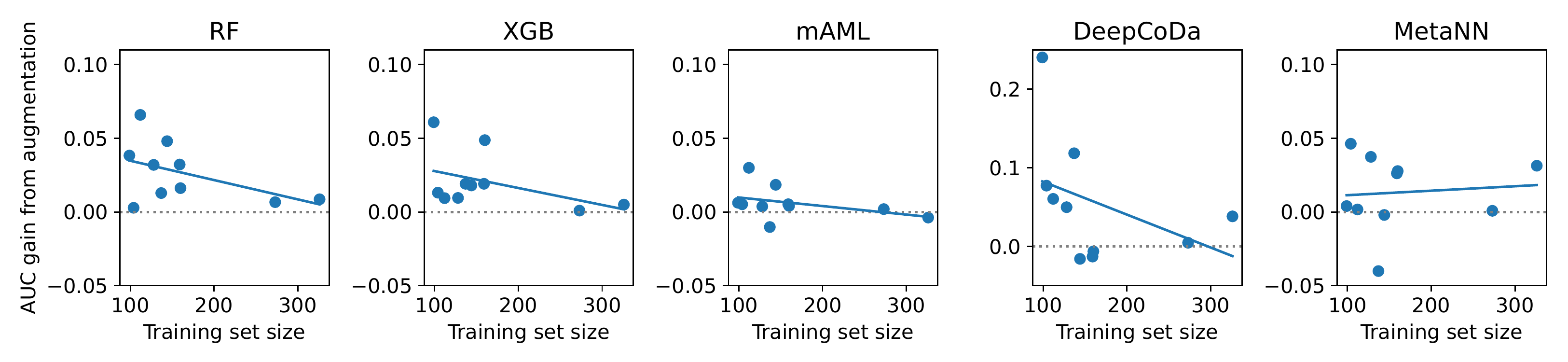}
  \caption{Difference in test AUC between models trained with Compositional CutMix, and those without, shown as a function of training set size.
  Each point represents one of our 12 benchmark learning tasks (note that tasks 3 and 5 are removed from the plot, since they already enjoyed 100\% classification accuracy prior to applying any data augmentation).
  The linear trendlines show a somewhat greater outperformance on the smaller datasets relative to the larger datasets.
  } \label{fig:abl}
\end{figure}


%
Figure \ref{fig:abl} shows the effect of data augmentation on test performance as a function of the number of samples in the training set, for Compositional CutMix.
We note that datasets with smaller sample sizes tended to benefit more from data augmentation.
This suggests that our methodology may prove beneficial to many microbiome research projects, where datasets with low-hundreds of samples are commonplace.
Figure \ref{fig:ablp} in Appendix \ref{app:exp} shows a similar scatterplot, as a function of the dataset dimensionality rather than sample size; the gains from augmentation appear consistent across lower- and higher-dimensional microbiome CoDa.

In addition, we evaluated the impact of our data augmentations on the expected calibration error (ECE) \cite{guo2017calibration} of our models.
Previous works have noted that increasingly flexible predictive models such as deep neural networks, while enabling greater predictive accuracy, tend to become overconfident in their predictions, degrading uncertainty quantification \cite{guo2017calibration}.
In Appendix \ref{app:exp}, we show the ECE obtained by our models, and we verify that our data augmentations do not hurt model calibration overall (in fact, some modest improvements are obtained).

\subsection{Contrastive representation learning} \label{sec:contrastive}

We have shown that our novel data augmentation strategies provide consistent performance gains across supervised training pipelines.
Next, we take our data augmentations one step further to provide the first application of contrastive representation learning to microbiome CoDa.

As a point of comparison, we use DeepMicro \cite{oh2020deepmicro}, a deep representation learning architecture tuned specifically for microbiome data.
DeepMicro trains a deep autoencoder to learn a low-dimensional representation of the microbial composition.
This learned representation is fed to a classification head downstream.
As per the author's implementation, the encoder has 2 hidden layers with 256 and 128 units respectively, and ReLU nonlinearities, and a 64-dimensional output.
The decoder architecture is a mirror image of the encoder.
The decoder output is of the same dimension as the encoder input, and the model is trained to minimize the mean squared error between the two, using the Adam optimizer with default parameters for 2,000 epochs.

To ensure a fair comparison, our contrastive model uses the same encoder architecture from DeepMicro.
However, we can discard the decoder layers; instead, the 64-dimensional latent representation is passed to a multilayer projection head that is trained jointly with the encoder using a contrastive loss, as is done in SimCLR \cite{chen2020simple}.
Our projection head contains one 32-dimensional hidden layer with ReLU activations, and a 16-dimensional projection output, which is normalized to 1 unit in L2 norm and passed to a temperature-scaled cross-entropy loss.
Intuitively, this loss function is designed to draw the representations of positive pairs of examples close together, and push those of negative pairs far apart.
Positive pairs refer to two synthetic samples generated by random data augmentations of the same training example; negative pairs refer to synthetic samples generated from different training examples.
In our implementation, at each training step we sample two Random Subcompositions from each training example, and compute the contrastive loss over all such pairs. 
In particular, our batch size corresponds to the entire training set, which is reasonable given the small sample sizes in our data.
We again used the Adam optimizer with default parameters for 2,000 epochs.

We evaluate the quality of our learned representations under two standard protocols.
First, linear evaluation, where the encoder weights are frozen and a linear classification head 
is trained using the supervised cross-entropy loss.
Second, finetuning, where the linear classification head is  
trained jointly with the encoder network. 
These evaluation protocols are applied for both DeepMicro and our contrastive model, in addition to a ``no pretraining'' control.
This control is a randomly initialized encoder network and linear head with the same architecture, trained only on the supervised objective.

The results are shown in Figure \ref{tab:contr}; the representations learned by our contrastive model consistently outperform those learned by DeepMicro, both in linear evaluation and finetuning.
Importantly, note that the comparison to DeepMicro is conservative, in the sense that we replicated the architecture and simply changed the pretraining objective from reconstruction error to contrastive loss; finetuning the encoder architecture itself under a contrastive objective would likely lead to further performance improvements.
We conclude that Random Subcompositions provide a valuable data augmentation for contrastive learning on the simplex.

Note that our other 2 augmentation strategies, Aitchison Mixup and Compositional CutMix, may also be used for contrastive learning.
However, these augmentations require paired training examples to generate each synthetic sample, and the implementation is therefore slightly more involved.
In particular, we generate each pair of positive samples by taking one pair of training examples, and drawing two random combinations (in the Aitchison Mixup or the Compositional CutMix sense) of that one pair.
The training examples are randomly partitioned into pairs at each epoch, and negative samples correspond to those that originate from disjoint pairs.
Note that we no longer require that the pairs are drawn from the same class for the contrastive objective.
The results are provided in Appendix \ref{app:exp} and show similarly strong performance to Table \ref{tab:contr}.

\begin{table}
  \caption{Representation learning performance for DeepMicro and contrastive learning (ours). For each task, we show the test AUC, averaged over 20 train/test bootstraps, under the linear evaluation protocol and finetuning. The representations obtained via contrastive learning consistently achieve higher AUC than those learned by DeepMicro. In addition, a randomly initialized encoder network of the same architecture is shown for comparison (No pretrain). Error bars are given in Appendix \ref{app:exp}.
}
  \label{tab:contr}
  \centering
  \footnotesize
    \vspace{3mm}
\begin{tabular}{c|c@{\hspace{1\tabcolsep}}c@{\hspace{1\tabcolsep}}c|c@{\hspace{1\tabcolsep}}c@{\hspace{1\tabcolsep}}c}
\toprule
   & \multicolumn{3}{c|}{Linear Evaluation } & \multicolumn{3}{c}{Finetuning} \\
Task & No pretrain & DeepMicro & Contrastive & No pretrain & DeepMicro & Contrastive \\
\midrule
1 & 0.59 & 0.68 & \textbf{0.76} & 0.72 & 0.75 & \textbf{0.77}\\
2 & 0.67 & 0.76 & \textbf{0.80} & 0.76 & 0.76 & \textbf{0.79}\\
3 & \textbf{1.00} & \textbf{1.00} & \textbf{1.00} & \textbf{1.00} & \textbf{1.00} & \textbf{1.00}\\
4 & 0.47 & 0.53 & \textbf{0.59} & 0.50 & 0.53 & \textbf{0.55}\\
5 & \textbf{1.00} & \textbf{1.00} & \textbf{1.00} & \textbf{1.00} & \textbf{1.00} & \textbf{1.00}\\
6 & 0.75 & 0.76 & \textbf{0.77} & \textbf{0.77} & 0.76 & \textbf{0.77}\\
7 & 0.59 & \textbf{0.68} & 0.63 & \textbf{0.59} & \textbf{0.59} & 0.57\\
8 & 0.65 & \textbf{0.66} & \textbf{0.66} & {0.66} & \textbf{0.68} & \textbf{0.68}\\
9 & 0.72 & 0.71 & \textbf{0.86} & 0.76 & 0.77 & \textbf{0.82}\\
10 & 0.53 & 0.58 & \textbf{0.62} & 0.60 & 0.61 & \textbf{0.63}\\
11 & 0.96 & \textbf{0.98} & \textbf{0.98} & \textbf{0.98} & \textbf{0.98} & \textbf{0.98}\\
12 & 0.66 & \textbf{0.68} & 0.66 & 0.62 & \textbf{0.64} & \textbf{0.64}\\
\midrule
Mean & 0.72 & 0.75 & \textbf{0.78} & 0.74 & 0.76 & \textbf{0.77}\\
\bottomrule
\end{tabular}
\end{table}

\section{Conclusion}

By combining ideas from data augmentation with the principles of CoDa, we have defined 3 novel augmentation strategies for CoDa: Aitchison Mixup, Random Subcompositions, and Compositional CutMix.
Our augmentations offer cheap performance gains across a wide range of benchmark datasets, advancing the state-of-the-art on standard classification benchmarks for colorectal cancer, type 2 diabetes, and Crohn's disease.
Our data augmentations can also be used to define an effective contrastive loss, which improves on existing representation learning frameworks for microbiome data.
Data augmentation and contrastive learning have enjoyed tremendous success in other application domains such as computer vision; our novel methodology aims to help spur similar developments in the fields of CoDa and the microbiome.


%

\section*{Acknowledgements}
We thank Richard Zemel and Samuel Lippl for helpful conversations.

\bibliographystyle{abbrv}
\bibliography{ref}

\newpage

\appendix

\section{Additional experimental results} \label{app:exp}

\begin{figure} [h]
  \centering
  \includegraphics[width=\textwidth]{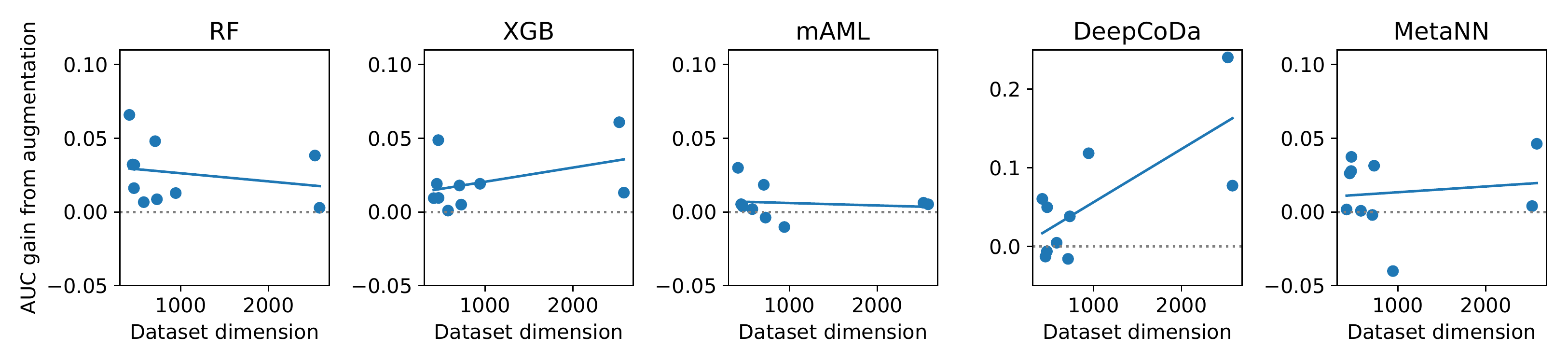}
  \caption{Similar to Figure \ref{fig:abl}. Difference in test AUC between models trained with Compositional CutMix, and those without, shown as a function of dataset dimension.
  } \label{fig:ablp}
\end{figure}

In Figure \ref{fig:ablp}, we show the gain in test AUC obtained from data augmentation (Compositional CutMix), plotted as a function of the dimensionality of the dataset.
Note that the performance gains from augmentation are broadly consistent across higher- and lower-dimensional datasets.
Tables \ref{tab:aitchece}, \ref{tab:randsubece}, and \ref{tab:subcompece} show the effect on expected calibration error (ECE) of adding data augmentation to our various models.
Notice that overall calibration error stays the same or improves slightly after incorporating data augmentation.
Tables \ref{tab:aitcherr}, \ref{tab:randsuberr}, and \ref{tab:subcomperr}, show error bars for Tables \ref{tab:aitch}, \ref{tab:randsub}, and \ref{tab:subcomp}, respectively.

In Tables \ref{tab:contram} and \ref{tab:contrcc}, we show the performance of our contrastive models defined with Aitchison Mixup and Compositional CutMix, respectively.
The latter performs equally strongly to Random Subcompositions, as shown in Section \ref{sec:contrastive}, the former performs noticeably worse, however still no worse than DeepMicro, thus demonstrating the robustness of our contrastive framework to multiple augmentation strategies.
Further gains may be obtained by combining augmentations in a contrastive loss, a direction which is left to future work.
Table \ref{tab:contr} shows error bars for Table \ref{tab:contrerr}, respectively.

\begin{table} [h]
  \caption{Augmentation performance for Aitchison Mixup. We show the test ECE, averaged over 20 train/test bootstraps, for each learning task and predictive model, trained with and without data augmentation.   }
  \label{tab:aitchece}
  \centering
  \footnotesize
    \vspace{3mm}
\begin{tabular}{c|c@{\hspace{1\tabcolsep}}c|c@{\hspace{1\tabcolsep}}c|c@{\hspace{1\tabcolsep}}c|c@{\hspace{1\tabcolsep}}c|c@{\hspace{1\tabcolsep}}c}
\toprule
Task & RF & Aug & XGB & Aug & mAML & Aug & DeepCoDa & Aug & MetaNN & Aug \\
\midrule
1 & {0.13} & {0.13} & 0.20 & {0.22} & {0.21} & 0.18 & 0.15 & {0.26} & 0.30 & {0.31}\\
2 & {0.14} & {0.14} & {0.19} & {0.19} & 0.15 & {0.18} & 0.18 & {0.21} & {0.31} & {0.31}\\
3 & {0.03} & 0.02 & {0.00} & {0.00} & {0.02} & {0.02} & {0.01} & 0.00 & {0.00} & {0.00}\\
4 & {0.10} & 0.08 & 0.27 & {0.29} & {0.23} & 0.19 & 0.15 & {0.27} & 0.46 & {0.47}\\
5 & {0.02} & 0.01 & {0.01} & 0.00 & {0.00} & {0.00} & {0.00} & {0.00} & {0.00} & {0.00}\\
6 & {0.11} & {0.11} & 0.13 & {0.15} & 0.10 & {0.11} & {0.14} & 0.12 & {0.26} & {0.26}\\
7 & {0.13} & {0.13} & 0.23 & {0.24} & 0.16 & {0.18} & 0.20 & {0.22} & 0.40 & {0.45}\\
8 & 0.08 & {0.12} & 0.24 & {0.25} & {0.20} & 0.19 & {0.43} & 0.23 & {0.38} & {0.38}\\
9 & {0.15} & 0.13 & {0.11} & {0.11} & {0.13} & {0.13} & {0.20} & 0.16 & {0.28} & 0.23\\
10 & {0.16} & 0.13 & 0.31 & {0.32} & {0.22} & {0.22} & {0.13} & 0.12 & 0.25 & {0.28}\\
11 & 0.07 & {0.08} & {0.04} & {0.04} & 0.07 & {0.08} & {0.07} & 0.06 & 0.05 & {0.06}\\
12 & {0.18} & 0.14 & {0.31} & 0.28 & 0.15 & {0.22} & 0.09 & {0.13} & 0.24 & {0.27}\\
\midrule
Mean & 0.11 & {0.10} & 0.17 & {0.17} & 0.14 & {0.14} & 0.15 & {0.15} & 0.24 & {0.24}\\
\bottomrule
\end{tabular}
\end{table}

\begin{table} [h]
  \caption{Augmentation performance for Random Subcompositions. We show the test ECE, averaged over 20 train/test bootstraps, for each learning task and predictive model, trained with and without data augmentation. }
  \label{tab:randsubece}
  \centering
  \footnotesize
    \vspace{3mm}
\begin{tabular}{c|c@{\hspace{1\tabcolsep}}c|c@{\hspace{1\tabcolsep}}c|c@{\hspace{1\tabcolsep}}c|c@{\hspace{1\tabcolsep}}c|c@{\hspace{1\tabcolsep}}c}
\toprule
Task & RF & Aug & XGB & Aug & mAML & Aug & DeepCoDa & Aug & MetaNN & Aug \\
\midrule
1 & {0.13} & {0.13} & 0.20 & {0.22} & {0.21} & {0.21} & 0.15 & {0.24} & {0.30} & 0.26\\
2 & {0.14} & 0.13 & {0.19} & 0.16 & {0.15} & {0.15} & {0.18} & 0.13 & {0.31} & 0.27\\
3 & {0.03} & 0.01 & {0.00} & {0.00} & {0.02} & {0.02} & {0.01} & 0.00 & {0.00} & {0.00}\\
4 & {0.10} & {0.10} & {0.27} & {0.27} & {0.23} & 0.21 & {0.15} & 0.12 & {0.46} & {0.46}\\
5 & {0.02} & 0.00 & {0.01} & 0.00 & 0.00 & {0.01} & {0.00} & {0.00} & {0.00} & {0.00}\\
6 & {0.11} & 0.10 & {0.13} & {0.13} & 0.10 & {0.11} & {0.14} & 0.13 & {0.26} & 0.24\\
7 & {0.13} & 0.11 & 0.23 & {0.24} & 0.16 & {0.17} & {0.20} & 0.12 & 0.40 & {0.44}\\
8 & 0.08 & {0.09} & 0.24 & {0.26} & {0.20} & 0.19 & {0.43} & 0.23 & {0.38} & 0.37\\
9 & {0.15} & 0.11 & 0.11 & {0.12} & {0.13} & {0.13} & {0.20} & 0.18 & {0.28} & 0.24\\
10 & {0.16} & {0.16} & {0.31} & 0.29 & {0.22} & {0.22} & {0.13} & 0.06 & {0.25} & 0.24\\
11 & 0.07 & {0.08} & {0.04} & {0.04} & 0.07 & {0.10} & {0.07} & 0.05 & {0.05} & {0.05}\\
12 & {0.18} & 0.17 & {0.31} & 0.27 & 0.15 & {0.17} & 0.09 & {0.10} & 0.24 & {0.25}\\
\midrule
Mean & 0.11 & {0.10} & 0.17 & {0.17} & 0.14 & {0.14} & 0.15 & {0.11} & 0.24 & {0.24}\\
\bottomrule
\end{tabular}
\end{table}

\begin{table} [h]
  \caption{Augmentation performance for Compositional CutMix. We show the test ECE, averaged over 20 train/test bootstraps, for each learning task and predictive model, trained with and without data augmentation. }
  \label{tab:subcompece}
  \centering
  \footnotesize
    \vspace{3mm}
\begin{tabular}{c|c@{\hspace{1\tabcolsep}}c|c@{\hspace{1\tabcolsep}}c|c@{\hspace{1\tabcolsep}}c|c@{\hspace{1\tabcolsep}}c|c@{\hspace{1\tabcolsep}}c}
\toprule
Task & RF & Aug & XGB & Aug & mAML & Aug & DeepCoDa & Aug & MetaNN & Aug \\
\midrule
1 & {0.13} & {0.13} & 0.20 & {0.22} & {0.21} & 0.19 & 0.15 & {0.24} & {0.30} & 0.29\\
2 & {0.14} & 0.13 & {0.19} & {0.19} & 0.15 & {0.17} & {0.18} & 0.17 & {0.31} & 0.29\\
3 & {0.03} & 0.01 & {0.00} & {0.00} & {0.02} & {0.02} & {0.01} & 0.00 & {0.00} & {0.00}\\
4 & {0.10} & {0.10} & {0.27} & {0.27} & {0.23} & 0.15 & 0.15 & {0.25} & 0.46 & {0.48}\\
5 & {0.02} & 0.01 & {0.01} & 0.00 & {0.00} & {0.00} & {0.00} & {0.00} & {0.00} & {0.00}\\
6 & {0.11} & 0.09 & 0.13 & {0.16} & 0.10 & {0.12} & {0.14} & 0.12 & {0.26} & 0.24\\
7 & 0.13 & {0.14} & 0.23 & {0.24} & {0.16} & {0.16} & {0.20} & 0.18 & 0.40 & {0.44}\\
8 & 0.08 & {0.11} & {0.24} & 0.23 & 0.20 & {0.22} & {0.43} & 0.27 & {0.38} & {0.38}\\
9 & {0.15} & 0.11 & {0.11} & 0.10 & {0.13} & 0.12 & {0.20} & 0.15 & {0.28} & 0.25\\
10 & {0.16} & {0.16} & {0.31} & {0.31} & {0.22} & 0.19 & {0.13} & {0.13} & 0.25 & {0.26}\\
11 & {0.07} & 0.06 & {0.04} & {0.04} & 0.07 & {0.10} & {0.07} & 0.05 & 0.05 & {0.06}\\
12 & {0.18} & {0.18} & {0.31} & 0.29 & 0.15 & {0.19} & 0.09 & {0.12} & 0.24 & {0.25}\\
\midrule
Mean & 0.11 & {0.10} & 0.17 & {0.17} & 0.14 & {0.14} & 0.15 & {0.14} & 0.24 & {0.24}\\
\bottomrule
\end{tabular}
\end{table}

\begin{table} [h]
  \caption{Similar to Table \ref{tab:contr}, but the contrastive model is now defined using Aitchison Mixup.}
  \label{tab:contram}
  \centering
  \footnotesize
    \vspace{3mm}
\begin{tabular}{c|c@{\hspace{1\tabcolsep}}c@{\hspace{1\tabcolsep}}c|c@{\hspace{1\tabcolsep}}c@{\hspace{1\tabcolsep}}c}
\toprule
   & \multicolumn{3}{c|}{Linear Evaluation } & \multicolumn{3}{c}{Finetuning} \\
Task & No pretrain & DeepMicro & Contrastive & No pretrain & DeepMicro & Contrastive \\
\midrule
1 & 0.59 & 0.68 & \textbf{0.72} & 0.72 & 0.75 & \textbf{0.77}\\
2 & 0.67 & 0.76 & \textbf{0.77} & 0.76 & 0.76 & \textbf{0.77}\\
3 & \textbf{1.00} & \textbf{1.00} & 0.96 & \textbf{1.00} & \textbf{1.00} & \textbf{1.00}\\
4 & 0.47 & \textbf{0.53} & \textbf{0.53} & 0.50 & \textbf{0.53} & \textbf{0.53}\\
5 & \textbf{1.00} & \textbf{1.00} & 0.99 & \textbf{1.00} & \textbf{1.00} & \textbf{1.00}\\
6 & 0.75 & \textbf{0.76} & \textbf{0.76} & 0.77 & 0.76 & \textbf{0.78}\\
7 & 0.59 & 0.68 & \textbf{0.75} & 0.59 & 0.59 & \textbf{0.65}\\
8 & 0.65 & \textbf{0.66} & 0.60 & 0.66 & \textbf{0.68} & 0.66\\
9 & 0.72 & 0.71 & \textbf{0.77} & 0.76 & 0.77 & \textbf{0.79}\\
10 & 0.53 & \textbf{0.58} & 0.54 & 0.60 & \textbf{0.61} & \textbf{0.61}\\
11 & 0.96 & \textbf{0.98} & 0.97 & \textbf{0.98} & \textbf{0.98} & 0.97\\
12 & 0.66 & \textbf{0.68} & 0.63 & 0.62 & \textbf{0.64} & 0.62\\
\midrule
Mean & 0.72 & 0.75 & {0.75} & 0.74 & 0.76 & {0.76}\\
\bottomrule
\end{tabular}
\end{table}

\begin{table} [h]
  \caption{Similar to Table \ref{tab:contr}, but the contrastive model is now defined using Compositional CutMix.}
  \label{tab:contrcc}
  \centering
  \footnotesize
    \vspace{3mm}
\begin{tabular}{c|c@{\hspace{1\tabcolsep}}c@{\hspace{1\tabcolsep}}c|c@{\hspace{1\tabcolsep}}c@{\hspace{1\tabcolsep}}c}
\toprule
   & \multicolumn{3}{c|}{Linear Evaluation } & \multicolumn{3}{c}{Finetuning} \\
Task & No pretrain & DeepMicro & Contrastive & No pretrain & DeepMicro & Contrastive \\
\midrule
1 & 0.59 & 0.68 & \textbf{0.76} & 0.72 & 0.75 & \textbf{0.76}\\
2 & 0.67 & 0.76 & \textbf{0.80} & 0.76 & 0.76 & \textbf{0.77}\\
3 & \textbf{1.00} & \textbf{1.00} & \textbf{1.00} & \textbf{1.00} & \textbf{1.00} & \textbf{1.00}\\
4 & 0.47 & 0.53 & \textbf{0.57} & 0.50 & 0.53 & \textbf{0.54}\\
5 & \textbf{1.00} & \textbf{1.00} & \textbf{1.00} & \textbf{1.00} & \textbf{1.00} & \textbf{1.00}\\
6 & 0.75 & 0.76 & \textbf{0.80} & 0.77 & 0.76 & \textbf{0.78}\\
7 & 0.59 & 0.68 & \textbf{0.72} & 0.59 & 0.59 & \textbf{0.61}\\
8 & 0.65 & \textbf{0.66} & \textbf{0.66} & 0.66 & \textbf{0.68} & \textbf{0.68}\\
9 & 0.72 & 0.71 & \textbf{0.85} & 0.76 & 0.77 & \textbf{0.81}\\
10 & 0.53 & 0.58 & \textbf{0.63} & 0.60 & 0.61 & \textbf{0.63}\\
11 & 0.96 & \textbf{0.98} & \textbf{0.98} & \textbf{0.98} & \textbf{0.98} & \textbf{0.98}\\
12 & 0.66 & \textbf{0.68} & 0.64 & 0.62 & \textbf{0.64} & 0.63\\
\midrule
Mean & 0.72 & 0.75 & \textbf{0.78} & 0.74 & 0.76 & \textbf{0.77}\\
\bottomrule
\end{tabular}
\end{table}


\begin{table} [h]
  \caption{Error bars for Table \ref{tab:aitch}.
  }
  \label{tab:aitcherr}
  \centering
  \footnotesize
    \vspace{3mm}
\begin{tabular}{c|c@{\hspace{1\tabcolsep}}c|c@{\hspace{1\tabcolsep}}c|c@{\hspace{1\tabcolsep}}c|c@{\hspace{1\tabcolsep}}c|c@{\hspace{1\tabcolsep}}c}
\toprule
Task & RF & Aug & XGB & Aug & mAML & Aug & DeepCoDa & Aug & MetaNN & Aug \\
\midrule
1 & {0.02} & 0.01 & {0.01} & {0.01} & {0.02} & {0.02} & {0.01} & {0.01} & {0.02} & {0.02}\\
2 & {0.02} & {0.02} & {0.02} & {0.02} & {0.02} & {0.02} & {0.02} & {0.02} & {0.02} & {0.02}\\
3 & {0.00} & {0.00} & {0.00} & {0.00} & {0.00} & {0.00} & {0.00} & {0.00} & {0.00} & {0.00}\\
4 & {0.02} & {0.02} & {0.02} & {0.02} & {0.02} & {0.02} & {0.02} & {0.02} & 0.01 & {0.02}\\
5 & {0.00} & {0.00} & {0.00} & {0.00} & {0.00} & {0.00} & {0.00} & {0.00} & {0.00} & {0.00}\\
6 & {0.01} & {0.01} & {0.01} & {0.01} & {0.01} & {0.01} & {0.01} & {0.01} & {0.01} & {0.01}\\
7 & {0.02} & {0.02} & {0.02} & {0.02} & {0.02} & {0.02} & {0.02} & {0.02} & {0.02} & {0.02}\\
8 & 0.02 & {0.03} & {0.02} & {0.02} & {0.03} & 0.02 & {0.02} & {0.02} & {0.03} & {0.03}\\
9 & {0.01} & {0.01} & {0.01} & {0.01} & {0.01} & {0.01} & {0.02} & {0.02} & {0.02} & {0.02}\\
10 & {0.02} & {0.02} & {0.02} & {0.02} & {0.02} & {0.02} & {0.02} & {0.02} & {0.02} & {0.02}\\
11 & {0.00} & {0.00} & {0.00} & {0.00} & {0.00} & {0.00} & {0.00} & {0.00} & {0.01} & {0.01}\\
12 & {0.02} & {0.02} & {0.02} & {0.02} & {0.02} & {0.02} & {0.02} & 0.01 & {0.02} & {0.02}\\
\midrule
Mean & 0.01 & {0.01} & 0.01 & {0.01} & 0.01 & {0.01} & 0.01 & {0.01} & 0.01 & {0.02}\\
\bottomrule
\end{tabular}
\end{table}

\begin{table} [h]
  \caption{Error bars for Table \ref{tab:randsub}. }
  \label{tab:randsuberr}
  \centering
  \footnotesize
    \vspace{3mm}
\begin{tabular}{c|c@{\hspace{1\tabcolsep}}c|c@{\hspace{1\tabcolsep}}c|c@{\hspace{1\tabcolsep}}c|c@{\hspace{1\tabcolsep}}c|c@{\hspace{1\tabcolsep}}c}
\toprule
Task & RF & Aug & XGB & Aug & mAML & Aug & DeepCoDa & Aug & MetaNN & Aug \\
\midrule
1 & {0.02} & 0.01 & {0.01} & {0.01} & {0.02} & {0.02} & {0.01} & {0.01} & {0.02} & {0.02}\\
2 & {0.02} & {0.02} & {0.02} & {0.02} & {0.02} & {0.02} & {0.02} & {0.02} & {0.02} & {0.02}\\
3 & {0.00} & {0.00} & {0.00} & {0.00} & {0.00} & {0.00} & {0.00} & {0.00} & {0.00} & {0.00}\\
4 & {0.02} & {0.02} & {0.02} & {0.02} & {0.02} & {0.02} & {0.02} & {0.02} & 0.01 & {0.02}\\
5 & {0.00} & {0.00} & {0.00} & {0.00} & {0.00} & {0.00} & {0.00} & {0.00} & {0.00} & {0.00}\\
6 & {0.01} & {0.01} & {0.01} & {0.01} & {0.01} & {0.01} & {0.01} & {0.01} & {0.01} & {0.01}\\
7 & {0.02} & {0.02} & {0.02} & {0.02} & {0.02} & {0.02} & {0.02} & {0.02} & {0.02} & {0.02}\\
8 & {0.02} & {0.02} & 0.02 & {0.03} & {0.03} & {0.03} & 0.02 & {0.03} & {0.03} & {0.03}\\
9 & {0.01} & {0.01} & {0.01} & {0.01} & {0.01} & {0.01} & {0.02} & {0.02} & {0.02} & {0.02}\\
10 & {0.02} & {0.02} & {0.02} & {0.02} & {0.02} & {0.02} & {0.02} & {0.02} & {0.02} & {0.02}\\
11 & {0.00} & {0.00} & {0.00} & {0.00} & {0.00} & {0.00} & {0.00} & {0.00} & {0.01} & {0.01}\\
12 & {0.02} & {0.02} & {0.02} & {0.02} & {0.02} & {0.02} & {0.02} & {0.02} & {0.02} & {0.02}\\
\midrule
Mean & 0.01 & {0.01} & 0.01 & {0.01} & 0.01 & {0.01} & 0.01 & {0.01} & 0.01 & {0.02}\\
\bottomrule
\end{tabular}
\end{table}

\begin{table} [h]
  \caption{Error bars for Table \ref{tab:subcomp}.
 }
  \label{tab:subcomperr}
  \centering
  \footnotesize
    \vspace{3mm}
\begin{tabular}{c|c@{\hspace{1\tabcolsep}}c|c@{\hspace{1\tabcolsep}}c|c@{\hspace{1\tabcolsep}}c|c@{\hspace{1\tabcolsep}}c|c@{\hspace{1\tabcolsep}}c}
\toprule
Task & RF & Aug & XGB & Aug & mAML & Aug & DeepCoDa & Aug & MetaNN & Aug \\
\midrule
1 & {0.02} & 0.01 & {0.01} & {0.01} & {0.02} & {0.02} & {0.01} & {0.01} & {0.02} & 0.01\\
2 & {0.02} & {0.02} & {0.02} & {0.02} & {0.02} & {0.02} & {0.02} & {0.02} & {0.02} & {0.02}\\
3 & {0.00} & {0.00} & {0.00} & {0.00} & {0.00} & {0.00} & {0.00} & {0.00} & {0.00} & {0.00}\\
4 & {0.02} & {0.02} & {0.02} & {0.02} & {0.02} & {0.02} & {0.02} & {0.02} & {0.01} & {0.01}\\
5 & {0.00} & {0.00} & {0.00} & {0.00} & {0.00} & {0.00} & {0.00} & {0.00} & {0.00} & {0.00}\\
6 & {0.01} & {0.01} & {0.01} & {0.01} & {0.01} & {0.01} & {0.01} & {0.01} & {0.01} & {0.01}\\
7 & {0.02} & {0.02} & {0.02} & {0.02} & {0.02} & {0.02} & {0.02} & {0.02} & {0.02} & {0.02}\\
8 & {0.02} & {0.02} & 0.02 & {0.03} & {0.03} & {0.03} & {0.02} & {0.02} & {0.03} & 0.02\\
9 & {0.01} & {0.01} & {0.01} & {0.01} & {0.01} & {0.01} & {0.02} & 0.01 & {0.02} & {0.02}\\
10 & {0.02} & {0.02} & {0.02} & {0.02} & {0.02} & {0.02} & {0.02} & {0.02} & {0.02} & {0.02}\\
11 & {0.00} & {0.00} & {0.00} & {0.00} & {0.00} & {0.00} & {0.00} & {0.00} & {0.01} & {0.01}\\
12 & {0.02} & {0.02} & {0.02} & {0.02} & {0.02} & {0.02} & {0.02} & 0.01 & {0.02} & {0.02}\\
\midrule
Mean & 0.01 & {0.01} & 0.01 & {0.01} & 0.01 & {0.01} & 0.01 & {0.01} & 0.01 & {0.01}\\
\bottomrule
\end{tabular}
\end{table}

\begin{table}
  \caption{Error bars for Table \ref{tab:contr}.
}
  \label{tab:contrerr}
  \centering
  \footnotesize
    \vspace{3mm}
\begin{tabular}{c|c@{\hspace{1\tabcolsep}}c@{\hspace{1\tabcolsep}}c|c@{\hspace{1\tabcolsep}}c@{\hspace{1\tabcolsep}}c}
\toprule
   & \multicolumn{3}{c|}{Linear Evaluation } & \multicolumn{3}{c}{Finetuning} \\
Task & No pretrain & DeepMicro & Contrastive & No pretrain & DeepMicro & Contrastive \\
\midrule
1 & {0.02} & {0.02} & 0.01 & {0.02} & {0.02} & 0.01\\
2 & {0.03} & 0.02 & 0.02 & {0.02} & {0.02} & {0.02}\\
3 & {0.00} & {0.00} & {0.00} & {0.00} & {0.00} & {0.00}\\
4 & {0.02} & {0.02} & {0.02} & 0.01 & 0.01 & {0.02}\\
5 & {0.00} & {0.00} & {0.00} & {0.00} & {0.00} & {0.00}\\
6 & {0.01} & {0.01} & {0.01} & {0.01} & {0.01} & {0.01}\\
7 & {0.02} & {0.02} & {0.02} & {0.02} & {0.02} & {0.02}\\
8 & {0.02} & {0.02} & {0.02} & {0.02} & {0.02} & {0.02}\\
9 & {0.02} & {0.02} & {0.02} & {0.02} & {0.02} & {0.02}\\
10 & 0.01 & {0.02} & {0.02} & {0.02} & {0.02} & {0.02}\\
11 & {0.00} & {0.00} & {0.00} & {0.00} & {0.00} & {0.00}\\
12 & {0.02} & 0.01 & 0.01 & {0.02} & {0.02} & {0.02}\\
\midrule
Mean & 0.01 & 0.01 & {0.01} & 0.01 & 0.01 & {0.01}\\
\bottomrule
\end{tabular}
\end{table}

\section{Other augmentation strategies} \label{app:otheraug}

We propose an additional augmentation strategy, \emph{Multinomial Resampling}, which can be thought of as a CoDa analogue of image blurring.
The goal of this augmentation is to inject noise across the coordinates of $\bold x$ in a principled manner.
Additive Gaussian noise is clearly ill-suited for CoDa, as it is not constrained to the simplex.
Instead, inspired by the recording mechanism for microbiome CoDa, we propose generating new datapoints from a multinomial distribution.
Note that high-throughput sequencing technologies record the microbial composition present in a specimen by subsampling the larger population.
The total number of reads in this subsample, known as the sequencing depth, is an artifact of the measurement process.
Assuming the subsample is small relative to the population, the multinomial distribution provides a crude approximation for this generative process, with each trial being drawn according to the true proportions in the underlying population.
In this way, we can sample new datapoints from a  multinomial distribution where the number of trials corresponds to the sequencing depth.
Namely, each new datapoint is generated as follows:
\begin{enumerate}
 \item Draw a training point $i$ uniformly at random.
 \item Draw $\bold{\tilde x}  \sim \text{Multinomial}(L_i, \bold x_i) $, where $L_i$ is the sequencing depth (a.k.a. library size). 
 \item Set $\bold x^{\text{aug}} = \bold{\tilde x} / ({\sum_{j=1}^p \tilde x_{j}}) $ and $y^{\text{aug}} = y_i$.
\end{enumerate}
Notice that the number of trials $L_i$ controls the noise level; as $L_i \to \infty$, $ \bold x^{\text{aug}} \to \bold x_i$.
When applied to CoDa that does not arise from high-throughput sequencing, $L_i$ can be specified arbitrarily, or treated as a hyperparameter.
In our microbiome datasets, the sequencing depth is typically on the order of $L \sim 10\,000$.

\end{document}